\documentclass[runningheads]{llncs}
\usepackage[utf8]{inputenc}
\usepackage{booktabs}
\usepackage{graphicx}
\usepackage{multirow}
\usepackage[misc]{ifsym}
\usepackage[utf8]{inputenc}
\usepackage{url}
\usepackage{xcolor}
\usepackage{lmodern}
\usepackage{caption}
\usepackage{floatrow}
\usepackage[outercaption]{sidecap}
\usepackage{amsmath}
\usepackage{cite}

\begin{document}

% ~~~~~~~~~~~~~~~~~~~~~ TITLE ~~~~~~~~~~~~~~~~~~~~~

\title{On the Replicability of Combining Word Embeddings and Retrieval Models}

% ~~~~~~~~~~~~~~~~~~~~~ AUTHORS ~~~~~~~~~~~~~~~~~~~~~

\author{Luca Papariello \and
Alexandros Bampoulidis \and
Mihai Lupu}

\institute{Research Studio Data Science, RSA FG \\ Vienna, Austria \\
    \email{\{name.surname\}@researchstudio.at}}

\maketitle

% ~~~~~~~~~~~~~~~~~~~~~ ABSTRACT ~~~~~~~~~~~~~~~~~~~~~

\begin{abstract}
We replicate recent experiments attempting to demonstrate an attractive hypothesis about the use of the Fisher kernel framework and mixture models for aggregating word embeddings towards document representations and the use of these representations in document classification, clustering, and retrieval. Specifically, the hypothesis was that the use of a mixture model of von Mises-Fisher (VMF) distributions instead of Gaussian distributions would be beneficial because of the focus on cosine distances of both VMF and the vector space model traditionally used in information retrieval. Previous experiments had validated this hypothesis. Our replication was not able to validate it, despite a large parameter scan space.
\end{abstract}

% ~~~~~~~~~~~~~~~~~~~~~ INTRODUCTION ~~~~~~~~~~~~~~~~~~~~~

\section{Introduction}

The last 5 years have seen proof that neural network-based word embedding models provide term representations that are a useful information source for a variety of tasks in natural language processing. In information retrieval (IR), ``traditional'' models remain a high baseline to beat, particularly when considering efficiency in addition to effectiveness~\cite{Hofstatter:2019a}. Combining the word embedding models with the traditional IR models is therefore very attractive and several papers have attempted to improve the baseline by adding in, in a more or less ad-hoc fashion, word-embedding information. Onal et al.~\cite{Onal:2018} summarized the various developments of the last half-decade in the field of neural IR and group the methods in two categories: \emph{aggregate} and \emph{learn}. The first one, also known as \emph{compositional distributional semantics}, starts from term representations and uses some function to combine them into a document representation (a simple example is a weighted sum). The second method uses the word embedding as a first layer of another neural network to output a document representation.

The advantage of the first type of methods is that they often distill down to a linear combination (perhaps via a kernel), from which an explanation about the representation of the document is easier to induce than from the neural network layers built on top of a word embedding. Recently, the issue of explainability in IR and recommendation is generating a renewed interest~\cite{Zhang:2019}.

In this sense, Zhang et al.~\cite{Zhang:2018d}  introduced a new model for combining high-dimensional vectors, using a mixture model of von Mises-Fisher (VMF)  instead of Gaussian distributions previously suggested by Clinchant and Perronnin~\cite{Clinchant:2013}. This is an attractive hypothesis because the Gaussian Mixture Model (GMM) works on Euclidean distance, while the mixture of von Mises-Fisher (moVMF) model works on cosine distances---the typical distance function in IR.

In the following sections, we set up to replicate the experiments described by Zhang et al.~\cite{Zhang:2018d}. They are grouped in three sets: classification, clustering, and information retrieval, and compare ``standard'' embedding methods with the novel moVMF representation.

% ~~~~~~~~~~~~~~~~~~~~~ EXPERIMENTS ~~~~~~~~~~~~~~~~~~~~~

\section{Experimental Setup}

In general, we follow the experimental setup of the original paper and, for lack of space, we do not repeat here many details, if they are clearly explained there.

\subsection{Datasets}  \label{sec:datasets}

All experiments are conducted on publicly available datasets and are briefly described here below.

\begin{description}
    \item[Classification.] Two subsets of the movie review dataset: (i) the subjectivity dataset (subj)~\cite{Pang_subj}; and (ii) the sentence polarity dataset (sent)~\cite{Pang_sent}.
    \item[Clustering.] The 20 Newsgroups dataset\footnote{\url{http://qwone.com/~jason/20Newsgroups/}} was used in the original paper, but the concrete version was not specified. We selected the ``bydate'' version, because it is, according to its creators, the most commonly used in the literature. It is also the version directly load-able in scikit-learn\footnote{\url{https://scikit-learn.org/stable/modules/generated/sklearn.datasets.fetch_20newsgroups.html}}, making it therefore more likely that the authors had used this version.
    \item[Retrieval.] The TREC Robust04 collection~\cite{Voorhees:2005}.
\end{description}

\subsection{Models}

The methods used to generate vectors for terms and documents are:

\begin{description}
    \item[TF-IDF.] The basic term frequency - inverse document frequency method~\cite{Harris_BOW}.
\end{description}
    \vspace{-0.25cm}
    \emph{Implemented in the scikit-learn library\footnote{\url{https://scikit-learn.org/stable/}}.}
\begin{description}
    \item[LSI.]  Latent Semantic Indexing~\cite{Deerwester_LSI}.
    \item[LDA.] Latent Dirichlet Allocation~\cite{Blei_LDA}.
    \item[cBoW.] Word2vec~\cite{Mikolov_w2v} in the  Continuous Bag-of-Word (cBow) architecture.
    \item[PV-DBOW/DM.] Paragraph vector (PV) is a document embedding algorithm that builds on Word2vec. We use here both its implementations: Distributed Bag-of-Words (PV-DBOW) and Distributed Memory (PV-DM)~\cite{Le:2014}.
\end{description}
\vspace{-0.25cm}
\emph{The LSI, LDA, cBoW, and PV implementations are available in the gensim library\footnote{\url{https://radimrehurek.com/gensim/}}.}
\begin{description}
    \item[Fisher Kernel (FK).] The FK framework offers the option to aggregate word embeddings to obtain fixed-length representations of documents. We use Fisher vectors (FV) based on (i) a Gaussian mixture model (FV-GMM) and (ii) a mixture of von Mises-Fisher distributions (FV-moVMF)~\cite{Banerjee_moVMF}.
    \end{description}

   We first fit (i) a GMM and (ii) a moVMF model on previously learnt continuous word embeddings. The fixed-length representation of a document $X$ containing $T$ words $w_i$---expressed as $X = \{ E_{w_1}, \ldots, E_{w_T} \}$, where $E_{w_i}$ is the word vector representation of word $w_i$---is then given by $\mathcal{G}^X = [\mathcal{G}_1^X, \ldots, \mathcal{G}_K^X]$, where $K$ is the number of mixture components. The vectors $\mathcal{G}_i^X$, having the dimension ($d$) of the word vectors $E_{w_i}$, are explicitly given by~\cite{Clinchant:2013, Zhang:2018d}:
\begin{equation}
        \textrm{(i)} \quad \mathcal{G}_i^X = \frac{1}{\sqrt{\omega_i}} \sum_{t=1}^T \gamma_t (i) \frac{x_t - \mu_i}{\sigma_i} , \quad \textrm{and} \quad \textrm{(ii)} \quad \mathcal{G}_i^X = \sum_{t=1}^T \frac{\gamma_t (i) x_t d}{\omega_i \kappa_i} ,
    \end{equation}
where $\omega_i$ are the mixture weights, $\gamma_t (i) = p(i | x_t )$ is the soft assignment of $x_t$ to (i) Gaussian and (ii) VMF distribution $i$, and $\sigma_i^2 = \textrm{diag}(\Sigma_i)$, with $\Sigma_i$ the covariance matrix of Gaussian $i$. In (i), $\sigma_i$ refers to the mean vector; in (ii) it indicates the mean direction and $\kappa_i$ is the concentration parameter.

We implement the FK-based algorithms by ourselves, with the help of the scikit-learn library for fitting a mixture of Gaussian models and of the Spherecluster package\footnote{\url{https://github.com/jasonlaska/spherecluster}} for fitting a mixture of von Mises-Fisher distributions to our data. The implementation details of each algorithm are described in what follows.

% ~~~~~~~~~~~~~~~~~~~~~ RESULTS ~~~~~~~~~~~~~~~~~~~~~

\section{Experimental Results}

Each of the following experiments is conceptually divided in three phases. First, text processing (e.g. tokenisation); second, creating  a fixed-length vector representation for every document; finally, the third phase is determined by the goal to be achieved, i.e. classification, clustering, and retrieval.

For the first phase the same pre-processing is applied to all datasets. In the original paper, this phase was only briefly described as tokenisation and stop-word removal. It is not given what tokeniser, linguistic filters (stemming, lemmatisation, etc.), or stop word list were used. Knowing that the gensim library was used, we took all standard parameters (see provided code\footnote{\url{https://rsagit.researchstudio.at/lpapariello/ecir_2020.git}}). Gensim however does not come with a pre-defined stopword list, and therefore, based on our own experience, we used the one provided in the NLTK library\footnote{\url{https://www.nltk.org}} for English.

For the second phase, transforming terms and documents to vectors, Zhang et al.~\cite{Zhang:2018d} specify that all trained models are 50 dimensional. We have additionally experimented with dimensionality 20 (used by Clinchant and Perronnin~\cite{Clinchant:2013} for clustering) and 100, as we hypothesized that 50 might be too low.
The TF-IDF model is 5000 dimensional (i.e. only the top 5000 terms based on their tf-idf value are used), while the Fischer-Kernel models are $15 \times d$ dimensional, where $d = \{ 20, 50, 100 \}$, as just explained. In what follows, $d$ refers to the dimensionality of LSI, LDA, cBow, and PV models.

The cBoW and PV models are trained using a default window size of 5, keeping both low and high-frequency terms, again following the setup of the original experiment. The LDA model is trained using a chunk size of 1000 documents and for a number of iterations over the corpus ranging from 20 to 100. For the FK methods, both fitting procedures (GMM and moVMF) are independently initialised 10 times and the best fitting model is kept.

For the third phase, parameters are explained in the following sections.

\subsection{Classification}

\begin{figure}[ht]
	\includegraphics[width=\linewidth]{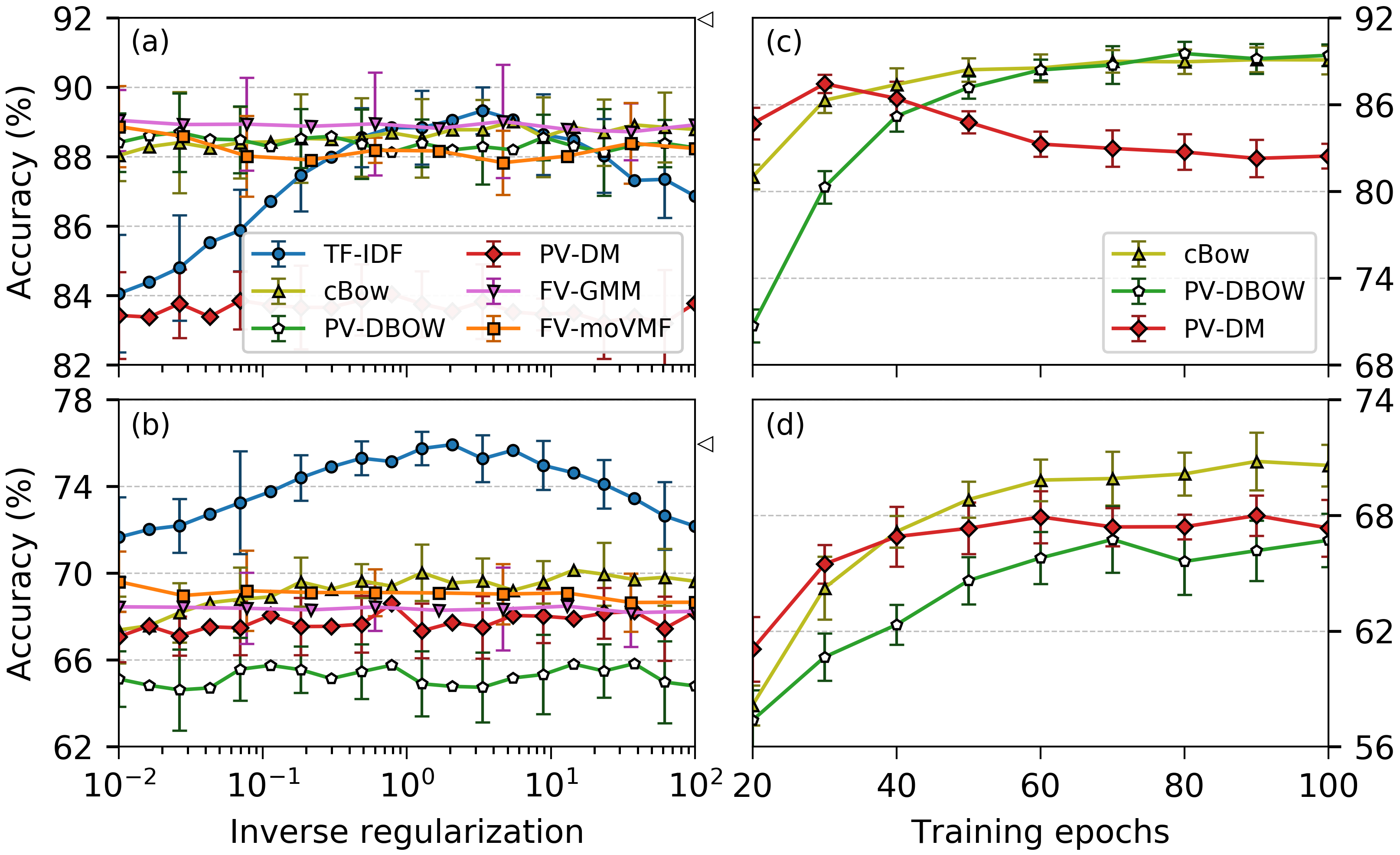}
	\caption{Results of classification experiments, for 50-dimensional feature vectors, on the \emph{subj} dataset [top panels, (a) and (c)] and \emph{sent} dataset [bottom panels, (b) and (d)]. LSI and LDA achieve low accuracy (see Table~\ref{tab:classif}) and are omitted here for visibility. The left panels [(a) and (b)] show the effect of (inverse) regularisation of the logistic regression classifier on the accuracy, while the right panels [(c) and (d)] display the effect of training for the learning algorithms. The two symbols on the right axis in panels (a) and (b) indicate the best (FV-moVMF) results reported in~\cite{Zhang:2018d}.}
    \label{fig:classif}
\end{figure}

Logistic regression is used for classification in Zhang et al., and therefore also used here.
The results of our experiments, for $d = 50$ and 100-dimensional feature vectors, are summarised in Table~\ref{tab:classif}. For all the methods, we perform a parameter scan of the (inverse) regularisation strength of the logistic regression classifier, as shown in Fig.~\ref{fig:classif}(a) and (b). Additionally, the learning algorithms are trained for a different number of epochs and the resulting classification accuracy assessed, cf. Fig.~\ref{fig:classif}(c) and (d).

\begin{SCtable}[][ht]
\small
\begin{tabular}{|l|c|c|c|c|}
  \hline
  \multirow{2}{*}{Model} & \multicolumn{2}{c|}{50-dim.} & \multicolumn{2}{c|}{100-dim.} \\ \cline{2-5}
   & Subj & Sent & Subj & Sent \\
  \hline
  TF-IDF & 89.3 $\pm$ 0.7 & 75.9 $\pm$ 1.3 & 89.3 $\pm$ 0.7 & 75.9 $\pm$ 1.3 \\
  LSI & 82.5 $\pm$ 1.0 & 63.2 $\pm$ 0.9 & 84.3 $\pm$ 1.0 & 65.4 $\pm$ 1.6 \\
  LDA & 62.2 $\pm$ 1.7 & 56.5 $\pm$ 1.6 & 62.1 $\pm$ 1.0 & 57.1 $\pm$ 1.5 \\
  cBow & 89.0 $\pm$ 1.1 & 70.1 $\pm$ 1.3 & 89.2 $\pm$ 1.6 & 71.4 $\pm$ 1.1 \\
  PV-DBOW & 88.7 $\pm$ 1.1 & 65.8 $\pm$ 1.4 & 89.4 $\pm$ 0.8 & 68.3 $\pm$ 1.3 \\
  PV-DM & 84.0 $\pm$ 1.5 & 68.6 $\pm$ 1.3 & 88.0 $\pm$ 0.9 & 70.3 $\pm$ 1.1 \\
  FV-GMM & 89.1 $\pm$ 0.9 & 68.5 $\pm$ 1.5 & 88.7 $\pm$ 0.9 & 72.4 $\pm$ 1.1 \\
  FV-moVMF & 89.5 $\pm$ 0.9 & 70.1 $\pm$ 0.9 & 88.7 $\pm$ 1.4 & 71.2 $\pm$ 1.5 \\
  \hline
\end{tabular}
\caption{Results of classification experiments on the \emph{subj} and \emph{sent} datasets. Shown are the mean accuracy and standard deviation, under 10-fold cross-validation, for optimally chosen hyperparameters (i.e. top values in Figure~\ref{fig:classif}). }
\label{tab:classif}
\end{SCtable}

Fig.~\ref{fig:classif}(a) indicates that cBow, FV-GMM, FV-moVMF, and the simple TF-IDF, when properly tuned, exhibit a very similar accuracy on \emph{subj}---the given confidence intervals do not indeed allow us to identify a single, best model. Surprisingly, TF-IDF outperforms all the others on the \emph{sent} dataset [Fig.~\ref{fig:classif}(b)]. Increasing the dimensionality of the feature vectors, from $d = 50$ to 100, has the effect of reducing the gap between TF-IDF and the rest of the models on the \emph{sent} dataset (see Table~\ref{tab:classif}).

\subsection{Clustering}

For clustering experiments, the obtained feature vectors are passed to the k-means algorithm. The results of our experiments, measured in terms of Adjusted Rand Index (ARI) and Normalized Mutual Information (NMI), are summarised in Table~\ref{tab:cluster}. We used both $d = 20$ and 50-dimensional feature vectors. Note that the evaluation of the clustering algorithms is based on the knowledge of the ground truth class assignments, available in the 20 Newsgroups dataset.

\begin{SCtable}[][h!]
\small
\begin{tabular}{|l|c|c|c|c|}
  \hline
  \multirow{2}{*}{Model} & \multicolumn{2}{c|}{20-dim.} & \multicolumn{2}{c|}{50-dim.} \\ \cline{2-5}
   & ARI & NMI & ARI & NMI \\
  \hline
  TF-IDF & 0.4 $\pm$ 0.1 & 5.6 $\pm$ 0.3 & 0.4 $\pm$ 0.1 & 5.6 $\pm$ 0.3 \\
  LSI & 0.6 $\pm$ 0.1 & 5.3 $\pm$ 0.8 & 0.5 $\pm$ 0.2 & 5.8 $\pm$ 0.3 \\
  LDA & 23.0 $\pm$ 0.7 & 39.5 $\pm$ 0.2 & 20.8 $\pm$ 1.1 & 40.2 $\pm$ 0.8 \\
  cBow & 31.2 $\pm$ 0.4 & 50.4 $\pm$ 0.4 & 31.5 $\pm$ 0.3 & 51.2 $\pm$ 0.3 \\
  PV-DBOW & 47.6 $\pm$ 1.2 & 63.4 $\pm$ 0.7 & 53.3 $\pm$ 1.3 & 66.1 $\pm$ 0.6 \\
  PV-DM & 16.8 $\pm$ 0.6 & 44.8 $\pm$ 0.3 & 30.1 $\pm$ 0.9 & 53.2 $\pm$ 0.5 \\
  FV-GMM & 1.4 $\pm$ 0.1 & 9.2 $\pm$ 0.8 & 1.0 $\pm$ 0.1 & 9.8 $\pm$ 1.3 \\
  FV-moVMF & 2.1 $\pm$ 0.2 & 13.3 $\pm$ 1.3 & 1.6 $\pm$ 0.2 & 14.9 $\pm$ 1.6 \\
  \hline
\end{tabular}
\caption{Results of clustering experiments (mean performance and standard deviation over 10 runs) in terms of Adjusted Rand Index (ARI) and Normalised Mutual Information (NMI).}
\label{tab:cluster}
\end{SCtable}

As opposed to classification, clustering experiments show a generous imbalance in performance and firmly speak in favour of PV-DBOW. Interestingly, TF-IDF, FV-GMM, and FV-moVMF, all providing high-dimensional document representations, have a low clustering effectiveness.

\subsection{Document Retrieval}

For these experiments, we extracted from every document of the test collection all the raw text, and preprocessed it as described in the beginning of this section.
The documents were indexed and retrieved for BM25 with the Lucene 8.2 search engine.
We experimented with three topic processing ways: (1) title only, (2) description only, and (3) title and description.
The third way produces the best results and closest to the ones reported by Zhang et. al~\cite{Zhang:2018d}, and hence are the only ones reported here.

An important aspect of BM25 is the fact that the variation of its parameters \textit{k\textsubscript{1}} and \textit{b} could bring significant improvement in performance, as reported by Lipani et. al \cite{lipani2015}.
Therefore, we performed a parameter scan for $k_1 \in [0, 3]$ and $b \in [0, 1]$ with a 0.05 step size for both parameters.
For every TREC topic, the scores of the top 1000 documents retrieved from BM25 were normalised to [0,1] with the min-max normalisation method, and were used in calculating the scores of the documents for the combined models~\cite{Zhang:2018d}.

The original results, those of our replication experiments with standard ($k_1=1.2$ and $b=0.75$) and best BM25 parameter values---measured in terms of Mean Average Precision (MAP) and Precision at 20 (P@20)---are outlined in Table~\ref{tab:retrieval}.

\begin{table}
\caption{Results of retrieval experiments with $95\%$ confidence intervals.}
\footnotesize{
\begin{tabular}{|c|c|c|c|c|c|c|}
\hline
\multirow{2}{*}{Model} & \multicolumn{2}{c|}{Zhang et. al} & \multicolumn{2}{c|}{Replicated} & \multicolumn{2}{c|}{\begin{tabular}[c]{@{}c@{}}Replicated Best BM25\end{tabular}} \\ \cline{2-7}
 & MAP & P@20 & MAP & P@20 & MAP & P@20 \\ \hline
BM25 & 24.10 & 33.70 & 22.80 $\pm$ 2.51 & 33.21 $\pm$ 2.95 & 22.94 $\pm$ 2.50 & 33.63 $\pm$ 2.99 \\ \hline
LSI & 3.40 & 3.90 & 0.39 $\pm$ 0.33 & 1.20 $\pm$ 0.74 & - & - \\ \hline
LDA & 4.70 & 5.60 & 0.37 $\pm$ 0.13 & 0.96 $\pm$ 0.37 & - & - \\ \hline
cBow & 7.20 & 11.10 & 3.84 $\pm$ 0.78 & 8.84 $\pm$ 1.33 & - & - \\ \hline
FV-GMM & 9.80 & 12.40 & 5.72 $\pm$ 1.14 & 11.24 $\pm$ 1.79 & - & - \\ \hline
FV-moVMF & 11.20 & 13.90 & 3.16 $\pm$ 0.60 & 7.71 $\pm$ 1.23 & - & - \\ \hline
BM25+LSI & 25.30 & 36.60 & 23.11 $\pm$ 2.54 & 33.29 $\pm$ 2.95 & 23.22 $\pm$ 2.53 & 33.73 $\pm$ 3.00 \\ \hline
BM25+LDA & 25.30 & 36.30 & 22.88 $\pm$ 2.52 & 33.37 $\pm$ 2.96 & 22.99 $\pm$ 2.53 & 33.96 $\pm$ 2.98 \\ \hline
BM25+cBow & 25.30 & 36.50 & 23.78 $\pm$ 2.55 & 34.22 $\pm$ 2.95 & 23.88 $\pm$ 2.55 & 34.72 $\pm$ 2.98 \\ \hline
BM25+FV-GMM & 25.40 & 36.30 & 23.51 $\pm$ 2.55 & 34.08 $\pm$ 2.93 & 23.68 $\pm$ 2.54 & 34.54 $\pm$ 2.96 \\ \hline
BM25+FV-moVMF & 25.60 & 36.70 & 23.45 $\pm$ 2.53 & 33.88 $\pm$ 2.95 & 23.58 $\pm$ 2.54 & 34.34 $\pm$ 2.96 \\ \hline
\end{tabular}
}
\label{tab:retrieval}
\end{table}

% ~~~~~~~~~~~~~~~~~~~~~ CONCLUSIONS ~~~~~~~~~~~~~~~~~~~~~

\section{Conclusions}

We replicated previously reported experiments that presented evidence that a new mixture model, based on von Mises-Fisher distributions, outperformed a series of other models in three tasks (classification, clustering, and retrieval---when combined with standard retrieval models).

Since the source code was not released in the original paper,  important implementation and formulation details were omitted, and the authors never replied to our request for information, a significant effort has been devoted to reverse engineer the experiments. In general, for none of the tasks were we able to confirm the conclusions of the previous experiments:  we do not have enough evidence to conclude that FV-moVMF outperforms the other methods. The situation is rather different when considering the effectiveness of these document representations for clustering purposes: we find indeed that the FV-moVMF significantly underperforms, contradicting previous conclusions.
In the case of retrieval, although Zhang et. al's proposed method (FV-moVMF) indeed boosts BM25, it does not outperform most of the other models it was compared to.

% ~~~~~~~~~~~~~~~~~~~~~ ACKNOWLEDGMENT ~~~~~~~~~~~~~~~~~~~~~

\paragraph*{Acknowledgments} Authors are partially supported by the H2020 Safe-DEED project (GA 825225).

% ~~~~~~~~~~~~~~~~~~~~~ BIBLIOGRAPHY ~~~~~~~~~~~~~~~~~~~~~

\bibliography{bibli}{}

\begin{thebibliography}{10}

\bibitem{Banerjee_moVMF}
A.~Banerjee, I.~S. Dhillon, J.~Ghosh, and S.~Sra.
\newblock Clustering on the unit hypersphere using von mises-fisher
  distributions.
\newblock {\em J. Mach. Learn. Res.}, 6:1345--1382, December 2005.

\bibitem{Blei_LDA}
D.~M. Blei, A.~Y. Ng, and M.~I. Jordan.
\newblock Latent dirichlet allocation.
\newblock {\em J. Mach. Learn. Res.}, 3:993--1022, March 2003.

\bibitem{Clinchant:2013}
St{\'e}phane Clinchant and Florent Perronnin.
\newblock Aggregating continuous word embeddings for information retrieval.
\newblock In {\em Proceedings of the workshop on continuous vector space models
  and their compositionality}, pages 100--109, 2013.

\bibitem{Deerwester_LSI}
S.~Deerwester, S.~T. Dumais, G.~W. Furnas, T.~K. Landauer, and R.~Harshman.
\newblock Indexing by latent semantic analysis.
\newblock {\em Journal of the American Society for Information Science},
  41(6):391--407, 1990.

\bibitem{Harris_BOW}
Z.~S. Harris.
\newblock Distributional structure.
\newblock {\em Word}, 10(2-3):146--162, 1954.

\bibitem{Hofstatter:2019a}
S.~Hofst{\"{a}}tter and A.~Hanbury.
\newblock Let's measure run time! extending the {IR} replicability
  infrastructure to include performance aspects.
\newblock In {\em Proceedings of the Open-Source {IR} Replicability Challenge
  co-located with 42nd International {ACM} {SIGIR} Conference on Research and
  Development in Information Retrieval, OSIRRC@SIGIR 2019, Paris, France, July
  25, 2019.}, pages 12--16, 2019.

\bibitem{Le:2014}
Quoc Le and Tomas Mikolov.
\newblock Distributed representations of sentences and documents.
\newblock In {\em International conference on machine learning}, pages
  1188--1196, 2014.

\bibitem{lipani2015}
Aldo Lipani, Mihai Lupu, Allan Hanbury, and Akiko Aizawa.
\newblock Verboseness fission for bm25 document length normalization.
\newblock In {\em Proceedings of the 2015 International Conference on The
  Theory of Information Retrieval}, pages 385--388. ACM, 2015.

\bibitem{Mikolov_w2v}
T.~Mikolov, K.~Chen, G.~Corrado, and J.~Dean.
\newblock Efficient estimation of word representations in vector space, 2013.

\bibitem{Onal:2018}
Kezban~Dilek Onal, Ye~Zhang, Ismail~Sengor Altingovde, Md~Mustafizur Rahman,
  Pinar Karagoz, Alex Braylan, Brandon Dang, Heng-Lu Chang, Henna Kim, Quinten
  McNamara, Aaron Angert, Edward Banner, Vivek Khetan, Tyler McDonnell,
  An~Thanh Nguyen, Dan Xu, Byron~C. Wallace, Maarten de~Rijke, and Matthew
  Lease.
\newblock Neural information retrieval: at the end of the early years.
\newblock {\em Information Retrieval Journal}, 21(2):111--182, Jun 2018.

\bibitem{Pang_subj}
B.~Pang and L.~Lee.
\newblock A sentimental education: Sentiment analysis using subjectivity
  summarization based on minimum cuts.
\newblock In {\em Proceedings of the 42nd Annual Meeting of the Association for
  Computational Linguistics ({ACL}-04)}, pages 271--278, 2004.

\bibitem{Pang_sent}
B.~Pang and L.~Lee.
\newblock Seeing stars: Exploiting class relationships for sentiment
  categorization with respect to rating scales.
\newblock In {\em Proceedings of the 43rd Annual Meeting of the Association for
  Computational Linguistics ({ACL}{'}05)}, pages 115--124. Association for
  Computational Linguistics, 2005.

\bibitem{Voorhees:2005}
Ellen~M. Voorhees.
\newblock The trec robust retrieval track.
\newblock {\em SIGIR Forum}, 39(1):11--20, June 2005.

\bibitem{Zhang:2018d}
R.~Zhang, J.~Guo, Y.~Lan, J.~Xu, and X.~Cheng.
\newblock Aggregating neural word embeddings for document representation.
\newblock In Gabriella Pasi, Benjamin Piwowarski, Leif Azzopardi, and Allan
  Hanbury, editors, {\em Advances in Information Retrieval}, pages 303--315.
  Springer International Publishing, 2018.

\bibitem{Zhang:2019}
Yongfeng Zhang, Yi~Zhang, Min Zhang, and Chirag Shah.
\newblock {EARS} 2019: The 2nd international workshop on explainable
  recommendation and search.
\newblock In {\em Proceedings of the 42nd International {ACM} {SIGIR}
  Conference on Research and Development in Information Retrieval, {SIGIR}
  2019, Paris, France, July 21-25, 2019.}, pages 1438--1440, 2019.

\end{thebibliography}
\bibliographystyle{plain}

\end{document}